\documentclass{article}

\usepackage{microtype}
\usepackage[dvipdfmx]{graphicx}
\usepackage{subfigure}
\usepackage{booktabs} 

\usepackage{hyperref}

\usepackage[accepted]{whi2018}

\usepackage{times}
\usepackage{amsthm,amssymb,amsmath}
\usepackage{url}

\theoremstyle{definition}
\newtheorem{theorem}{Theorem}  [section]

\newtheorem{problem}[theorem] {Problem}

\numberwithin{equation}{section}

\DeclareMathOperator*{\argmax}{argmax}
\DeclareMathOperator*{\argmin}{argmin}

\icmltitlerunning{Maximally Invariant Data Perturbation as Explanation}

\begin{document}

\twocolumn[
\icmltitle{Maximally Invariant Data Perturbation as Explanation}



\icmlsetsymbol{equal}{*}

\begin{icmlauthorlist}
\icmlauthor{Satoshi Hara}{1}
\icmlauthor{Kouichi Ikeno}{1}
\icmlauthor{Tasuku Soma}{2}
\icmlauthor{Takanori Maehara}{3}
\end{icmlauthorlist}

\icmlaffiliation{1}{Osaka University, Osaka, Japan}
\icmlaffiliation{2}{The University of Tokyo, Tokyo, Japan}
\icmlaffiliation{3}{RIKEN AIP, Tokyo, Japan} 

\icmlcorrespondingauthor{Satoshi Hara}{satohara@ar.sanken.osaka-u.ac.jp}

\icmlkeywords{Interpretability, Data Perturbation, Linear Programming}

\vskip 0.3in
]



\printAffiliationsAndNotice{}  

\begin{abstract}
While several feature scoring methods are proposed to explain the output of complex machine learning models, most of them lack formal mathematical definitions.
In this study, we propose a novel definition of the feature score using the maximally invariant data perturbation, which is inspired from the idea of adversarial example.
In adversarial example, one seeks the smallest data perturbation that changes the model's output.
In our proposed approach, we consider the opposite: we seek the maximally invariant data perturbation that does not change the model's output.
In this way, we can identify important input features as the ones with small allowable data perturbations.
To find the maximally invariant data perturbation, we formulate the problem as linear programming.
The experiment on the image classification with VGG16 shows that the proposed method could identify relevant parts of the images effectively.
\end{abstract}

\section{Introduction}
\label{sec:intro}

Complex machine learning models like deep learning have attained significant performance improvements in several fields such as image recognition, speech recognition, and natural language processing.
However, complex models are usually less interpretable compared to the simple models such as the linear models and the rule models.
Recently, this black-box nature of complex machine learning models turn out to be problematic in some domains like medical diagnosis~\cite{ong2014diabetes}, judicial decisions~\cite{jordan2015effect}, and education~\cite{lakkaraju2015machine}, where the users request an explanation why the model made a certain decision.
It is therefore essential in those fields to provide an explanation for each model's output.

A popular explanation method for complex models is \textit{feature scoring}~\cite{erhan2009visualizing,springenberg2014striving,bach2015pixel,sundararajan2017axiomatic,smilkov2017smoothgrad,shrikumar2017learning}.
It scores the features relevant to model's output with large values, while scoring the irrelevant features with small values.
The users can then focus on high score features as an explanation of the output.

Although feature scoring methods are found to be useful, most of them lack formal mathematical definitions except for the axiomatic approach of \citet{sundararajan2017axiomatic}: most scores are defined by their computation algorithms themselves.
This means that there are no well-defined ground truth that we want to obtain.
To push the entire field forward, we need to formalize the feature scoring problem to be solved.
In this way, we can discuss whether the definitions are appropriate, and we can inspect which part of the algorithms are causing problems, which will lead to further methodological improvements.

The contributions of this paper are twofold.
First, we propose a novel formulation of the feature scoring problem inspired from the idea of adversarial example~\cite{szegedy2013intriguing}.
In adversarial example, one seeks the smallest data perturbation that changes the model's output.
In the proposed problem formulation, we consider the opposite: we seek the maximal data perturbation that does not change the model's output.
The intuition behind this approach is as follows.
If the feature is relevant to the output, it cannot change a lot to keep the output unchanged.
On the other hand, if the feature is not relevant to the output, it can vary arbitrary.
Therefore, by finding the maximal data perturbation that does not change the output, we can identify important input features as the ones with small allowable data perturbations.

The second contribution is the development of a computation method to solve the proposed problem.
To find the maximal data perturbation, we formulate the problem as a semi-infinite programming.
We then show that the problem can be approximated as linear programming, which can be solved efficiently using existing solvers.
We also present some practically useful extensions.

\paragraph{Settings}
In this paper, we consider the classification model $f$ for $k$ categories that returns an output $y \in \mathbb{R}^k$ for a given input $x \in \mathbb{R}^d$, i.e., $y = f(x)$.
The classification result is determined as $c = \argmax_j y_j$ where $y_j = f_j(x)$ is the $j$-th element of the output.
We assume that the model $f$ is differentiable with respect to the input $x$: the target models therefore include linear models, kernel models (with differentiable kernels), and deep learning.
We assume that the model $f$ and the target input $x$ to be explained are given and fixed.

\section{Problem Formulation}
\label{sec:problem}

In this section, we present our proposed problem formulation for feature scoring, inspired from the idea of adversarial example.
We first review the problem definition of the adversarial example, then present our problem formulation.

\subsection{Problem Definition of Adversarial Example}
In adversarial example, one seeks the minimum data perturbation that changes the model's output.
The problem is formulated as follows~\cite{szegedy2013intriguing}:
\begin{align}
	\hat{r} = \argmin_{r \in \mathbb{R}^d} \|r\|, \; {\rm s.t.} \; c \neq \argmax_j f_j(x + r) ,
	\label{eq:adv}
\end{align}
where $c = \argmax_j f_j(x)$ is the predicted class label for the given input $x$.
The solution $\hat{r}$ is the minimum data perturbation that leads to a different output.
The value $\|\hat{r}\|$ is usually referred as the robustness of the model $f$ because the model's output is invariant for all perturbation within this radius, i.e., for all $r \in \mathbb{R}^d$ with $\| r \| \le \| \hat{r} \|$, $c = \argmax_j f(x_j + r)$.

\subsection{Proposed Problem Formulation}

We propose to measure the relevance of features by the invariance with respect to the perturbation. 

To begin with, we introduce \emph{invariant perturbation set}.
A set $R \subseteq \mathbb{R}^d$ is an invariant perturbation set if the model's output is invariant for all $r \in R$. 
We can interpret that the problem~\eqref{eq:adv} seeks the largest ball-shaped invariant perturbation set of the problem.

In our study, for ease of computation, we restrict our attention to a box-shaped invariant perturbation set $R(u, v) = [-u_1, v_1] \times \cdots \times [-u_d, v_d]$ for parameters $u, v \in \mathbb{R}^d_+$. 
From this definition of $R(u, v)$, the invariant perturbation of each feature $x_i$ is $u_i + v_i$.
Here, it is important to note that, if the invariant perturbation $u_i + v_i$ is small, the change of the feature $x_i$ can highly impacts the output, which indicates that the feature $x_i$ is relevant to the output.
On the other hand, if $u_{i'} + v_{i'}$ is large, the feature $x_{i'}$ only has a minor impact to the output, and thus it is less relevant to the output.

We now define the \emph{maximal} invariant perturbation set to measure the relevance of features.
By measuring the size of $R(u, v)$ by $\sum_i (u_i + v_i)$, we consider the largest set $R(u, v)$ as the solution to the following optimization problem.
\begin{problem}[Maximal Invariant Perturbation]
	\label{prob:proposed}
	Find the invariant perturbation set $R(\hat{u}, \hat{v})$, where
	\begin{align}
		\begin{split}
		\hat{u}, \hat{v} &= \argmax_{u, v \in \mathbb{R}_+^d} \sum_{i=1}^d (u_i + v_i), \\
		& {\rm s.t.} \; \textstyle{c = \argmax_j f_j(x + r), \forall r \in R(u, v)}.
		\label{eq:problem}
		\end{split}
	\end{align}
\end{problem}
We note that the problem (\ref{eq:problem}) is a semi-infinite programming~\cite{hettich1993semi} that contains infinitely many constraints parametrized by $r$.

\section{Proposed Method}
\label{sec:method}

We now turn to our proposed method to solve the problem (\ref{eq:problem}).
The difficulty is that the constraint $c = \argmax_j f_j(x + r)$ tends to be complex when the model $f$ is highly nonlinear, which is almost the case in practice.
In the proposed method, we adopt a simple approximation of the constraint so that the problem to be tractable.

\subsection{Problem Approximation as Linear Programming}
In the proposed method, we approximate the problem (\ref{eq:problem}) as linear programming, which can then be solved using the state-of-the-art solvers such as CPLEX and Gurobi.

The basic idea of the proposed method is to refine the problem (\ref{eq:problem}) into a tractable formulation by using the first-order Taylor expansion:
\begin{align}
	f_j(x+ r) \approx f_j(x) + \nabla f_j(x)^\top r.
	\label{eq:taylor}
\end{align}
Because this approximation is valid only in the neighborhood of $x$, we restrict the perturbation $r$ to be sufficiently small.
Specifically, we restrict $r$ as $|r_i| \le \delta$ for sufficiently small $\delta > 0$, which is equivalent to upper bounding $u_i$ and $v_i$ as $u_i, v_i \le \delta$.

With the approximation (\ref{eq:taylor}), we can rewrite the constraint $c = \argmax_j f_j(x + r)$ as linear inequalities.
Because it is equivalent to $f_c(x+r) \ge f_j(x + r), \forall j \neq c$, the constraint can be approximated as
\begin{align}
	f_c(x) + \nabla f_c(x)^\top r \ge f_j(x) + \nabla f_j(x)^\top r ,
\end{align}
for any $j \neq c$.
By reorganizing the inequality, we obtain
\begin{align}
	g_j \ge h_j^\top r,
	\label{eq:ineq}
\end{align}
where $g_j := f_c(x) - f_j(x)$ and $h_j := \nabla f_j(x) - \nabla f_c(x)$. 

We now show that we can remove the dependency to $r$ in the constraint (\ref{eq:ineq}), and the problem (\ref{eq:problem}) can be reformulated into a linear programming.
Recall that $h_j^\top r = \sum_{i} h_{ji} r_i$ and $r_i \in [-u_i, v_i]$.
Then, the maximum of the term $h_{ji} r_i$ is equal to $h_{ji}v_i$ if $h_{ji} \ge 0$ and $-h_{ji}u_i$ otherwise.
Therefore, the constraint $g_j \ge h_j^\top r$ for the \emph{worst} $r \in R(u, v)$ can be expressed as
\begin{align}
	g_j \ge \sum_{i: h_{ji} \ge 0} h_{ji} v_i - \sum_{i': h_{ji'} < 0} h_{ji'} u_{i'} ,
\end{align}
which no longer depends on $r$.
The overall problem is then formulated as the following linear programming:
\begin{align}\label{eq:lp}
\begin{split}
	 \max_{u, v} & \sum_{i=1}^d (u_i + v_i),  \\
	 {\rm s.t.}\; 0 & \le u_i, v_i \le \delta, \forall i,  \\
	 \;\; g_j & \ge \sum_{i: h_{ji} \ge 0} h_{ji} v_i - \sum_{i': h_{ji'} < 0} h_{ji'} u_{i'}, \forall j \neq c.
\end{split}
\end{align}

\subsection{Useful Extensions}

Here, we present some practically useful extensions of the formulation (\ref{eq:lp}).

\paragraph{Soft-Constraint}
The inequality constraints in the problem (\ref{eq:lp}) can be too restrictive to enhance relevant features when there exists some similar categories (e.g., different kinds of dogs).
This is because, if some features are relevant to both of those categories, the inequality can suppress the invariant perturbation of these features to be small to prevent from changing between the similar categories.
To avoid this unfavorable suppressing, we propose replacing the inequality constraints with soft-constraints.
Specifically, we introduce a new parameter $w \ge 0$, and replace the inequality constraints with the following:
\begin{align}
	g_j \ge \sum_{i: h_{ji} \ge 0} h_{ji} v_i - \sum_{i': h_{ji'} < 0} h_{ji'} u_{i'} - w.
\end{align}
This extension requires the constraints to hold only up to the parameter $w$, which allows some constraints to be violated.
To optimize the parameter $w$, we introduce an additional penalty term in the objective function as $\sum_{i=1}^d (u_i + v_i) - \lambda w$, 
where $\lambda > 0$ is a penalty parameter that is determined by the user.
We note that the resulting problem is still a linear programming even under the soft-constraint setting.

\paragraph{Parameter-Sharing}
In image recognition tasks, because of the spatial smoothness of the images, it is less likely that the neighboring pixels have significantly different invariant perturbations (except the boundary of the objects).
To take this prior knowledge into account, we can extend the problem (\ref{eq:lp}) so that some features (e.g., neighboring pixels in the image) to share the same parameters $u_i$ and $v_i$.
Suppose that we have partitioned features to $M$ disjoint subsets $I_1, I_2, \ldots, I_M \subseteq [d] := \{1, 2, \ldots, d\}$ such that $I_m \cap I_{m'} = \emptyset$ and $\cup_m I_m = [d]$.
We then assume that the parameters are shared within each partition as $u_i = u_{i'} = u_m, v_i = v_{i'} = v_m$ for $\forall i, i' \in I_m$.
We can then reformulate the problem (\ref{eq:lp}) (with soft-constraint) with the parameter sharing as
\begin{align}\label{eq:lpshare}
	\begin{split}
		\max_{u, v, w \ge 0} & \sum_{m=1}^M (u_m + v_m) - \lambda w, \\
		{\rm s.t.} \; 0 & \le u_m, v_m \le \delta, \forall m, \\
		\; g_j & \ge \sum_{m=1}^M \left( h_{jm}^+ v_m - h_{jm}^- u_{m} \right) - w, \forall j \neq c ,
	\end{split}
\end{align}
where $h_{jm}^+ := \sum_{i \in I_m: h_{ji} \ge 0} h_{ji}$ and $h_{jm}^- := \sum_{i' \in I_m: h_{ji'} < 0} h_{ji'}$.

\paragraph{Smoothing}
\citet{smilkov2017smoothgrad} have shown that perturbing the input with small noises and then averaging the feature scores over perturbed inputs can smoothen the feature score and reduce the noises.
Here, we apply this idea to our method.
Specifically, we use the perturbed input $x + n$ with a small noise $n \in \mathbb{R}^d$ for Taylor expansion (\ref{eq:taylor}) as
\begin{align}
	f_j(x + r) \approx f_j(x + n) + \nabla f_j(x + n)^\top (r - n) .
	\label{eq:taylorn}
\end{align}
With the expression (\ref{eq:taylorn}), we can construct additional linear inequalities to the problem (\ref{eq:lp}):
\begin{align}
	g_j^{n} + h_j^\top n \ge \sum_{i: h_{ji}^{n} \ge 0} h_{ji}^{n} v_i - \sum_{i': h_{ji'}^{n} < 0} h_{ji'}^{n} u_{i'} ,
	\label{eq:ineqn}
\end{align}
where $g_j^{n} := f_c(x+n) - f_j(x+n)$ and $h_j^{n} := \nabla f_j(x+n) - \nabla f_c(x+n)$.
By using the additional constraints, we can gain more information of the model $f$ compared to the original formulation (\ref{eq:lp}).
We note that this smoothing extension can be used together with the soft-constraint and the parameter-sharing extensions.

\section{Experiment}
\label{sec:experiment}

\subsection{Experimental Setup}

In the experiment, we used the pre-trained VGG16~\cite{simonyan2014very} as the model $f$ distributed at the Tensorflow repository.
The VGG16 model takes the image of size $(224, 224, 3)$ as the input, and returns the probability for $k=1,000$ categories.
As the target data to be explained, we used COCO-animal dataset\footnote{\url{cs231n.stanford.edu/coco-animals.zip}} which contains images of eight animals.
From the obtained dataset, we used $200$ images in the validation set for our experiment.

As the proposed method, we used the formulation (\ref{eq:lpshare}) with the soft-constraint and the parameter-sharing.
For the parameter sharing, for each of RGB channel, we split the image of size $(224, 224)$ into non-overlapping $8 \times 8$ patches.
We then share the parameters within each patch.
The number of parameters in $u$ and $v$ are therefore $28 \times 28 \times 3 = 2352$ each.
We set the upper bound as $\delta = 0.1$ and the penalty as $\lambda = 2 \times 2352 \times 10^{-4}$.
The resulting score for the $m$th patch is computed as $2\delta - u_m - v_m$, which gets large when the allowable perturbations $u_m$ and $v_m$ are small.

We also adopted the smoothed version of the proposed method.
Specifically, we added the inequality constraint (\ref{eq:ineqn}) to the problem (\ref{eq:lpshare}) for nine different small perturbations $n$ generated from the independent Gaussian distribution with mean zero and standard deviation 0.05.

As the baseline methods, we used Gradient~\cite{erhan2009visualizing}, GuidedBP~\cite{springenberg2014striving}, SmoothGrad~\cite{smilkov2017smoothgrad}, IntGrad~\cite{sundararajan2017axiomatic}, LRP~\cite{bach2015pixel}, DeepLIFT~\cite{shrikumar2017learning}, and Occlusion.
For computing Gradient, GuidedBP, SmoothGrad and IntGrad, we used \texttt{saliency}\footnote{\url{https://github.com/PAIR-code/saliency}} with default settings, and for computing LRP, DeepLIFT, and Occlusion, we used \texttt{DeepExplain}\footnote{\url{https://github.com/marcoancona/DeepExplain}} where we set the mask size for Occlusion as $(12, 12, 3)$.
We note that, all the methods including the proposed methods return the feature scores of size $(224, 224, 3)$.

\subsection{Result}

We evaluated the efficacy of the proposed methods by partly masking images.
Even if we mask some parts of the images, the model's classification result will kept unchanged as long as relevant parts of the images remain unmasked.
To evaluate the performance of each scoring method, we masked the images as follows.
First, we flip pixels with scores smaller than the $\tau$\% quantile to $0.5$ (i.e., we replace the selected pixels with gray pixels\footnote{We also conducted experiments by replacing with zero (black) or one (white). The results were similar, and thus omitted.}).
We then observe the ratio of the images with the classification result changes within the 200 images: the result changes in less images indicate that the feature scoring methods successfully identified relevant parts of the images.

\begin{figure}[t]
	\centering
	\includegraphics[width=0.46\textwidth]{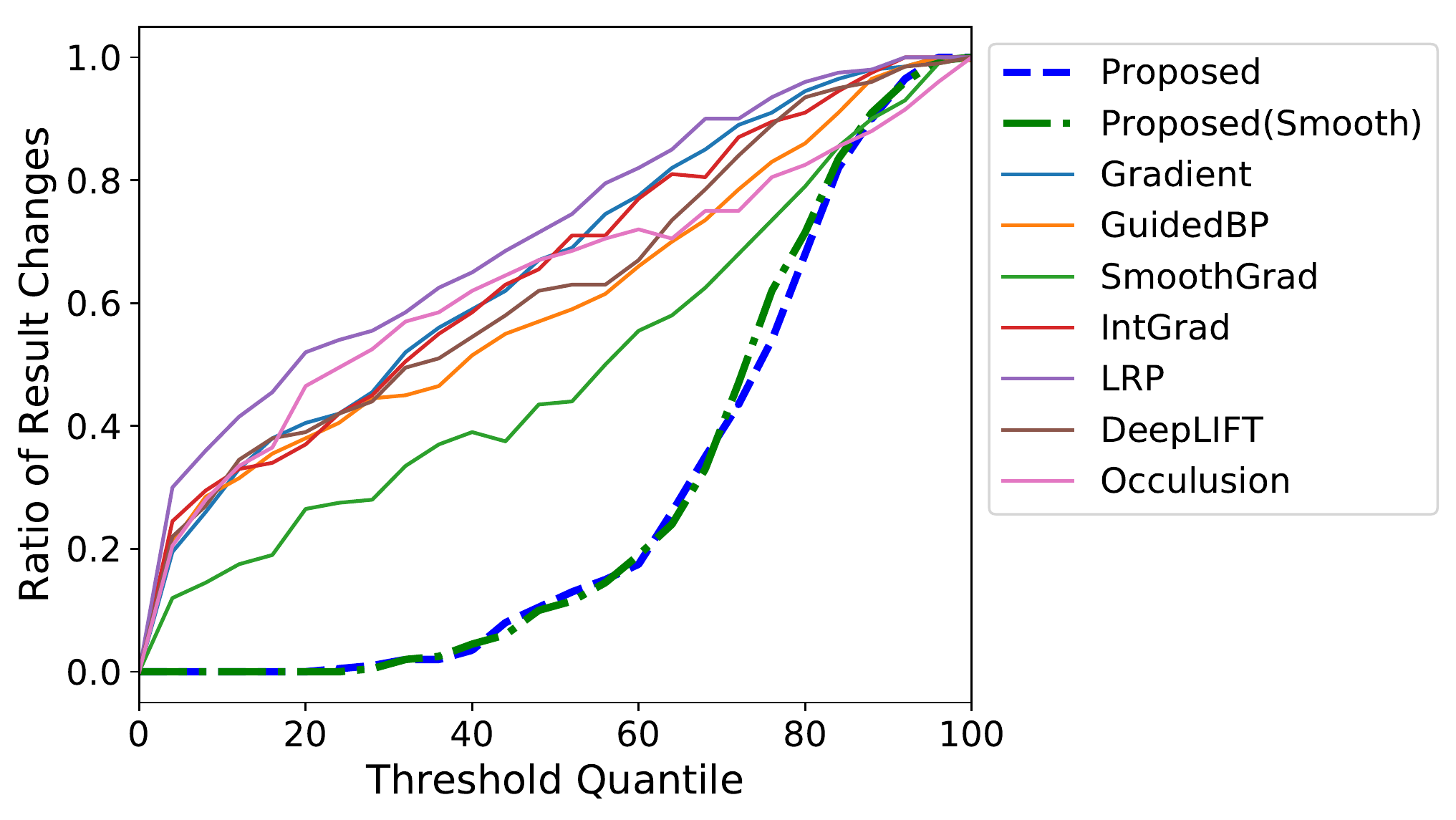}
	\caption{The ratio of the classification result changes when low score pixels are flipped to gray.}
	\label{fig:resvgg}
	\vspace{2pt}
	\centering
	\includegraphics[width=0.49\textwidth]{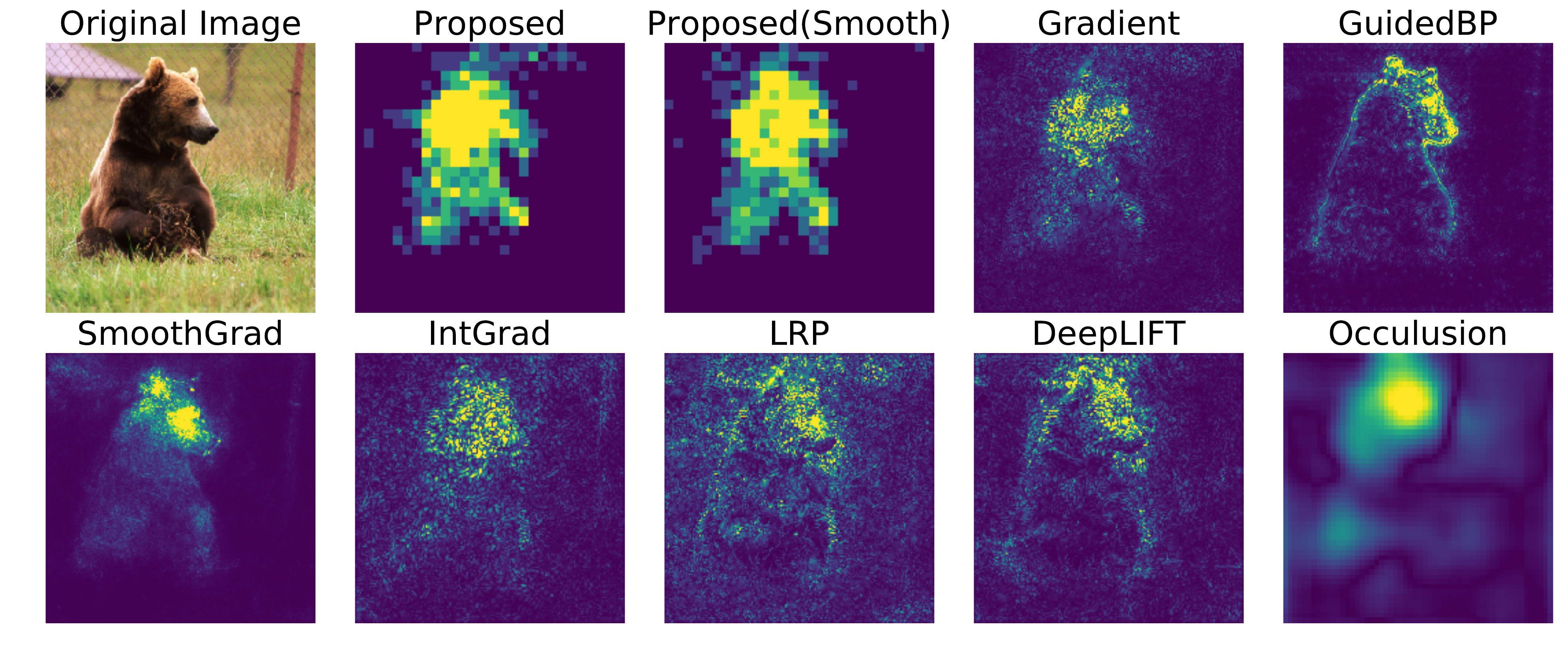}
	\includegraphics[width=0.49\textwidth]{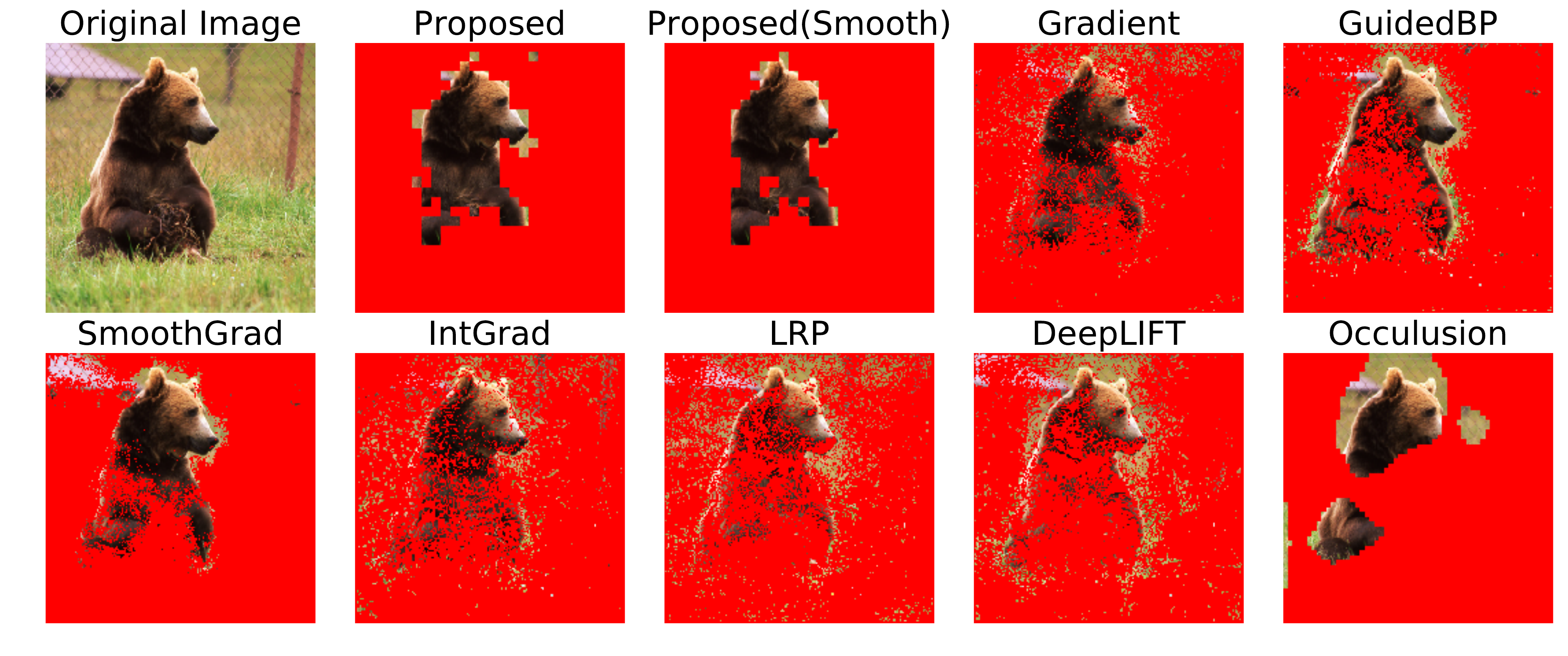}
	\caption{Example images: the computed scores as heatmap (top); the filtered images with pixels with scores lower than the 70\% quantiles are filtered out with red (bottom).}
	\label{fig:example}
\end{figure}

\figurename~\ref{fig:resvgg} shows the result when we varied the threshold quantile from $\tau = 0$ to $\tau = 100$.
It is clear that the proposed methods are resistant to the masking: the changes on the classification results are kept small even if 50\% of the pixels are masked.
This indicates that the proposed method successfully identified relevant parts of the images.

\figurename~\ref{fig:example} shows an example of the computed scores with each method.
The figures show that the proposed methods attained high S/N ratios thanks to the parameter sharing while the scores of the existing method tend to be noisy.
It is also interesting to see that the proposed method indicated that the entire body of the bear are relevant, while the other methods highlighted only the head of the bear.
We conjecture that this difference comes from the definition of the proposed score: the proposed score judged that even the changes on the bear body can induce the class changes.

\section{Conclusion}
\label{sec:conclusion}

In this study, we proposed a new definition of the feature score using the maximally invariant data perturbation, which is inspired from the idea of adversarial example.
We also proposed to formulate the problem as linear programming using the first-order Taylor expansion.
The experimental result on the image classification with VGG16 shows that the proposed method could identify relevant parts of the images effectively.

\section*{Acknowledgement}
This work was supported by JSPS KAKENHI Grant Number JP18K18106.

\bibliography{main}
\bibliographystyle{icml2018}

\end{document}